\documentclass[sigconf, anonymous=false]{acmart}
\acmSubmissionID{259}

\AtBeginDocument{%
  \providecommand\BibTeX{{%
    \normalfont B\kern-0.5em{\scshape i\kern-0.25em b}\kern-0.8em\TeX}}}

\setcopyright{acmcopyright}
\copyrightyear{2018}
\acmYear{2018}
\acmDOI{XXXXXXX.XXXXXXX}

\usepackage{subfig}
\usepackage{pbox}
\usepackage{fancyhdr}

\begin{document}
\settopmatter{printacmref=false}
\renewcommand\footnotetextcopyrightpermission[1]{}
\title{DensePose From WiFi}
\captionsetup[subfloat]{farskip=0pt,captionskip=1pt}


\author{Jiaqi Geng}
\email{jiaqigen@andrew.cmu.edu}
\affiliation{%
  \institution{Carnegie Mellon University}
  \city{Pittsburgh}
  \state{PA}
  \country{USA}
}

\author{Dong Huang}
\email{donghuang@cmu.edu}
\affiliation{%
  \institution{Carnegie Mellon University}
  \city{Pittsburgh}
  \state{PA}
  \country{USA}
}

\author{Fernando De la Torre}
\email{ftorre@cs.cmu.edu}
\affiliation{%
  \institution{Carnegie Mellon University}
  \city{Pittsburgh}
  \state{PA}
  \country{USA}
}

\renewcommand{\shortauthors}{Anonymous authors, et al.}


\begin{abstract}
Advances in computer vision and machine learning techniques have led to significant development in 2D and 3D human pose estimation from RGB cameras, LiDAR, and radars. However, human pose estimation from images is adversely affected by occlusion and lighting, which are common in many scenarios of interest. Radar and LiDAR technologies, on the other hand, need specialized hardware that is expensive and power-intensive. Furthermore, placing these sensors in non-public areas raises significant privacy concerns. 

To address these limitations, recent research has explored the use of WiFi antennas (1D sensors) for body segmentation and key-point body detection. This paper further expands on the use of the WiFi signal in combination with deep learning architectures, commonly used in computer vision, to estimate dense human pose correspondence. We developed a deep neural network that maps the phase and amplitude of WiFi signals to UV coordinates within 24 human regions. The results of the study reveal that our model can estimate the dense pose of multiple subjects, with comparable performance to image-based approaches, by utilizing WiFi signals as the only input. This paves the way for low-cost, broadly accessible, and privacy-preserving algorithms for human sensing.  

\end{abstract}




\maketitle
\pagestyle{plain}
\pagenumbering{gobble}

\begin{figure*}[htp]
 \centering
 \includegraphics[scale=0.35]{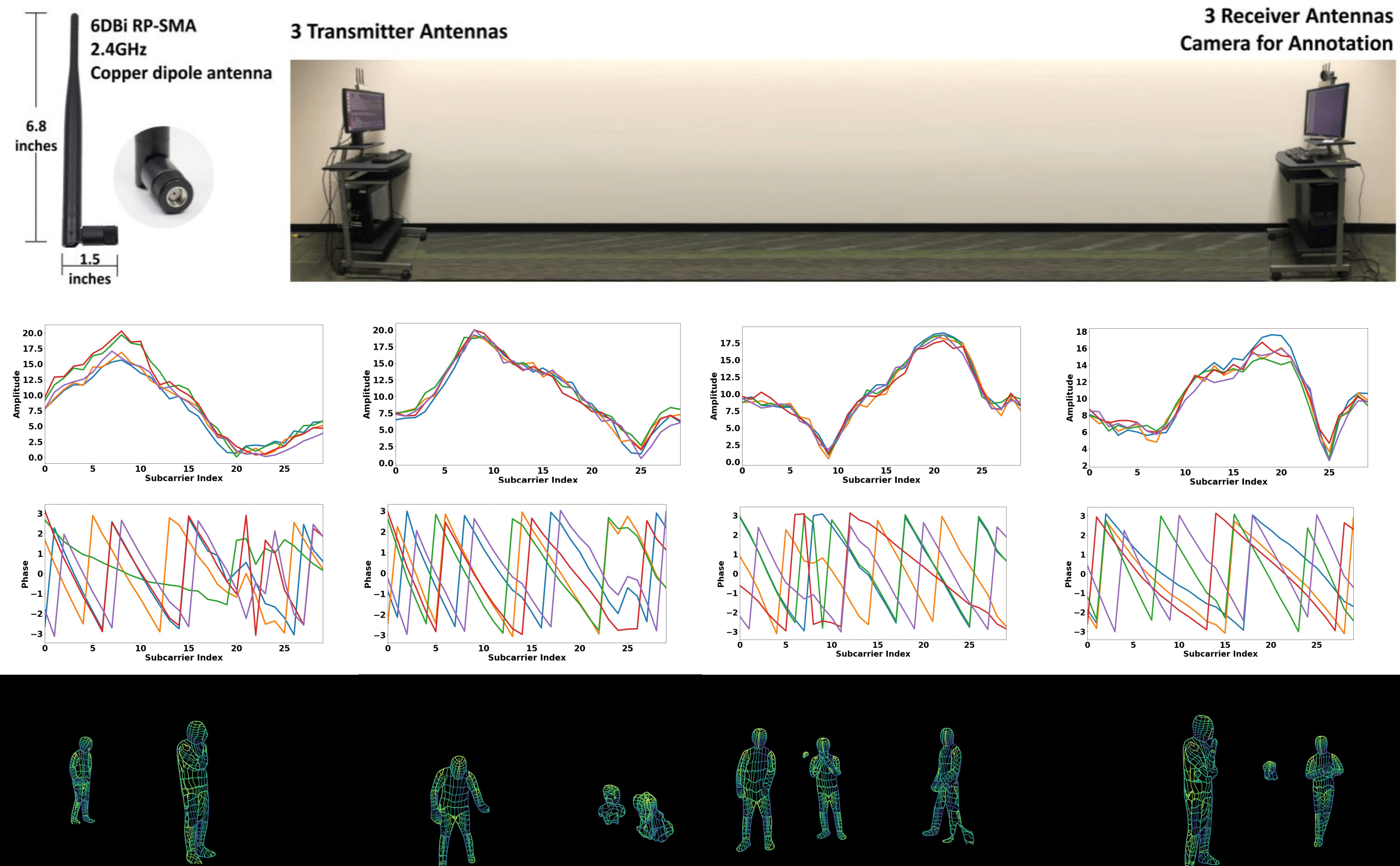}
 \caption{The first row illustrates the hardware setup. The second and third rows are the clips of amplitude and phase of the input WiFi signal. The fourth row contains the dense pose estimation of our algorithm from {\em only} the WiFi signal.}
 \label{Fig_intro}
\end{figure*}

\section{Introduction}

Much progress has been made in human pose estimation using 2D \cite{DensePose,convolutional-pose,ROMP,vibe,pifu,openpose} and 3D \cite{lidar-3d,voxnet} sensors in the last few years (e.g., RGB sensors, LiDARs, radars), fueled by applications in autonomous driving and augmented reality. These traditional sensors, however, are constrained by both technical and practical considerations. LiDAR and radar sensors are frequently seen as being out of reach for the average household or small business due to their high cost. For example, the medium price of one of the most common COTS LiDAR, Intel L515, is around 700 dollars, and the prices for ordinary radar detectors range from 200 dollars to 600 dollars. In addition, these sensors are too power-consuming for daily and household use. As for RGB cameras, narrow field of view and poor lighting conditions, such as glare and darkness, can have a severe impact on camera-based approaches. Occlusion is another obstacle that prevents the camera-based model from generating reasonable pose predictions in images. This is especially worrisome for indoors scenarios, where furniture typically occludes people. 

More importantly, privacy concerns prevent the use of these technologies in non-public places. For instance, most people are uncomfortable with having cameras recording them in their homes, and in certain areas (such as the bathroom) it will not be feasible to install them. These issues are particularly critical in healthcare applications, that are increasingly shifting from clinics to homes, where people are being monitored with the help of cameras and other sensors. It is important to resolve the aforementioned problems in order to better assist the aging population, which is the most susceptible (especially during COVID) and has a growing demand to keep them living independently at home.  

We believe that WiFi signals can serve as a ubiquitous substitute for RGB images for human sensing in certain instances. Illumination and occlusion have little effect on WiFi-based solutions used for interior monitoring. In addition, they protect individuals' privacy and the required equipment can be bought at a reasonable price. In fact, most households in developed countries already have WiFi at home,  and this technology may be scaled to monitor the well-being of elder people or just identify suspicious behaviors at home. 

The issue we are trying to solve is depicted in Fig. 1 (first row). Given three WiFi transmitters and three aligned receivers, can we detect and recover dense human pose correspondence in cluttered scenarios with multiple people (Fig. 1 fourth row). It should be noted that many WiFi routers, such as TP-Link AC1750, come with 3 antennas, so our method only requires 2 of these routers. Each of these router is around 30 dollars, which means our entire setup is still way cheaper than LiDAR and radar systems. 
Many factors make this a difficult task to solve.  First of all, WiFi-based perception\cite{Person-in-wifi,wi2vi} is based on the Channel-state-information (CSI) that represents the ratio between the transmitted signal wave and the received signal wave. The CSIs are complex decimal sequences that do not have spatial correspondence to spatial locations, such as the image pixels.  Secondly, classic techniques rely on accurate measurement of time-of-fly and angle-of-arrival of the signal between the transmitter and receiver \cite{Decimeter_Level_WiFi_Localization,ultra-wideband}. These techniques only locate the object's center; moreover, the localization accuracy is only around $0.5$ meters due to the random phase shift allowed by the IEEE 802.11n/ac WiFi communication standard and potential interference with electronic devices under similar frequency range such as microwave oven and cellphones.

To address these issues, we derive inspiration from recent proposed deep learning architectures in computer vision, and propose a neural network architecture that can perform dense pose estimation from WiFi. Fig~\ref{Fig_intro} (bottom row) illustrates how our algorithm is able to estimate dense pose using {\em only} WiFi signal in scenarios with occlusion and multiple people.

\section{Related Work}
This section briefly describes existing work on dense estimation from images and human sensing from WiFi. 

Our research aims to conduct dense pose estimation via WiFi. In computer vision, the subject of dense pose estimation from pictures and video has received a lot of attention \cite{DensePose,ContinuousSurfaceEmbeddings,bristow2015dense,zhou2016learning}. This task consists of finding the dense correspondence between image pixels and the dense vertices indexes of a 3D human body model. The pioneering work of G{\"{u}}ler et al. \cite{DensePose} mapped human images to dense correspondences of a human mesh model using deep networks. DensePose is based on instance segmentation architectures such as Mark-RCNN~\cite{Mask-RCNN}, and predicts body-wise UV maps for each pixel, where UV maps are flattened representations of 3d geometry, with coordinate points usually corresponding to the vertices of a 3d dimensional object. In this work, we borrow the same architecture as DensePose~\cite{DensePose}; however, our input will not be an image or video, but we use 1D WiFi signals to recover the dense correspondence. 

Recently, there have been many extensions of DensePose proposed, especially in 3D human reconstruction with dense body parts \cite{DenseRac,tex2shape,densebody,3d-from-dense-parts}. Shapovalov et al.'s \cite{DensePose3D} work focused on lifting dense pose surface maps to 3D human models without 3D supervision. Their network demonstrates that the dense correspondence alone (without using full 2D RGB images) contains sufficient information to generate posed 3D human body. Compared to previous works on reconstructing 3D humans with sparse 2D keypoints, DensePose annotations are much denser and provide information about the 3D surface instead of 2D body joints.  

While there is a extensive literature on detection \cite{region_proposal,yolov3}, tracking \cite{posetrack,simple_tracking}, 
and dense pose estimation \cite{DensePose,ContinuousSurfaceEmbeddings} from images and videos, human pose estimation
from WiFi or radar is a relatively unexplored problem. At this point, it is important to differentiate the current work on radar-based systems and WiFi.  The work of Adib et.al. \cite{radar} proposed a Frequency Modulated Continuous Wave (FMCW) radar system (broad bandwidth from 5.56GHz to 7.25GHz) for indoor human localization. A limitation of this system is the specialized hardware for synchronizing the transmission, refraction, and reflection to compute the Time-of-Flight (ToF). The system reached a resolution of 8.8 cm on body localization. In the following work \cite{RF-capture}, they improved the system by focusing on a moving person and generated a rough single-person outline with depth maps. Recently, they applied deep learning approaches to do fine-grained human pose estimation using a similar system, named RF-Pose \cite{RF-Pose}. These systems do not work under the IEEE 802.11n/ac WiFi communication standard (40MHz bandwidth centered at 2.4GHz). They rely on additional high-frequency and high-bandwidth electromagnetic fields, which need specialized technology not available to the general public. Recently, significant improvements have been made to radar-based human sensing systems. mmMesh \cite{xue2021mmmesh} generates 3D human mesh from commercially portable millimeter-wave devices. This system can accurately localize the vertices on the human mesh with an average error of 2.47 cm. However, mmMesh does not work well with occlusions since high-frequency radio waves cannot penetrate objects. 
 
Unlike the above radar systems, the WiFi-based solution~\cite{Person-in-wifi,wi2vi} used off-the-shelf WiFi adapters and 3dB omnidirectional antennas. The signal propagate as the IEEE 802.11n/ac WiFi data packages transmitting between antennas, which does not introduce additional interference.  However, WiFi-based person localization using the traditional time-of-flight (ToF) method is limited by its wavelength and signal-to-noise ratio. Most existing approaches only conduct center mass localization~\cite{BASRI2016,Soltanaghaei2018} and single-person action classification~\cite{Sheng2020,WangSong2019}.  Recently, 
Fei Wang et.al.~\cite{Wanghuang2019} demonstrated that it is possible to detect $17$ 2D body joints and perform 2D semantic body segmentation mask using only WiFi signals. In this work, we go beyond~\cite{Wanghuang2019} by estimating dense body pose, with much more accuracy than the 0.5m that the WiFi signal can provide theoretically.
Our dense posture outputs push above WiFi's signal constraint in body localization, 
paving the road for complete dense 2D and possibly 3D human body perception through WiFi. To achieve this, instead of directly training a randomly initialized WiFi-based model, we explored rich supervision information to improve both the performance and training efficiency, such as utilizing the CSI phase, adding keypoint detection branch, and transfer learning from the image-based model.

\section{Methods}
Our approach produces UV coordinates of the human body surface from WiFi signals using three components:
first, the raw CSI signals are cleaned by amplitude and phase sanitization. Then, a two-branch encoder-decoder network performs domain translation from sanitized CSI samples to 2D feature maps that resemble images. The 2D features are then fed to a modified DensePose-RCNN architecture \cite{DensePose} to estimate the UV map, a representation of the dense correspondence between 2D and 3D humans. To improve the training of our WiFi-input network, we conduct transfer learning, where we minimize the differences between the multi-level feature maps produced by images and those produced by WiFi signals before training our main network. 
 
 \begin{figure}[!htb]
\centering
\subfloat[\centering Layout of WiFi devices and human bodies]{{\includegraphics[scale=0.32]{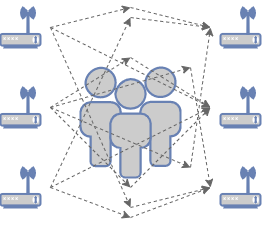}}}\ %
\subfloat[\centering The $3\times 3$ dimensions of the CSI tensor] {{\includegraphics[scale=0.2]{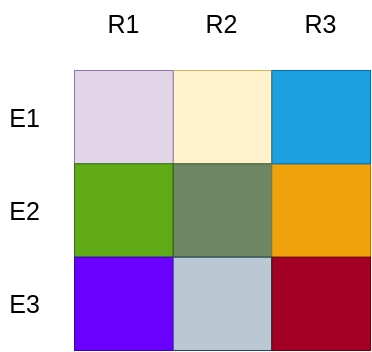} }}%
\caption{CSI samples from Wifi. (a) the layout of WiFi devices and human bodies, and (b) the $3 \times 3$ tensor dimension corresponds to the $3 \times 3$ transmitter-receiver antenna pairs. For instance, $E1$ denotes the first emitter and $R1$ denotes the first receiver, etc. By incorporating the 5 consecutive complex-valued CSI samples (100 samples/second) under 30 subcarrier frequencies, the two input tensors to our network are a $150 \times 3 \times 3$ amplitude tensor and a $150 \times 3 \times 3$ phase tensor.} %
\label{fig:CSIrepresenation}
\end{figure}
 
The raw CSI data are sampled in 100Hz as complex values over 30 subcarrier frequencies (linearly spaced within 2.4GHz$\pm$20MHz) transmitting among 3 emitter antennas and 3 reception antennas (see Figure~\ref{fig:CSIrepresenation}). Each CSI sample contains a $3 \times 3$ real integer matrix and a $3 \times 3$ imaginary integer matrix. The inputs of our network contained 5 consecutive CSI samples under 30 frequencies, which are organized in a $150 \times 3 \times 3$ amplitude tensor and a $150 \times 3 \times 3$ phase tensor respectively. Our network outputs include a $17 \times 56 \times 56$ tensor of keypoint heatmaps (one $56 \times 56$ map for each of the 17 kepoints) and a $25 \times 112 \times 112$ tensor of UV maps (one $112 \times 112$ map for each of the 24 body parts with one additional map for background). 
 
 \subsection{Phase Sanitization}

\begin{figure*}[!htb]
\centering
\subfloat[\centering Original CSI Amplitude]{{\includegraphics[scale=0.35]{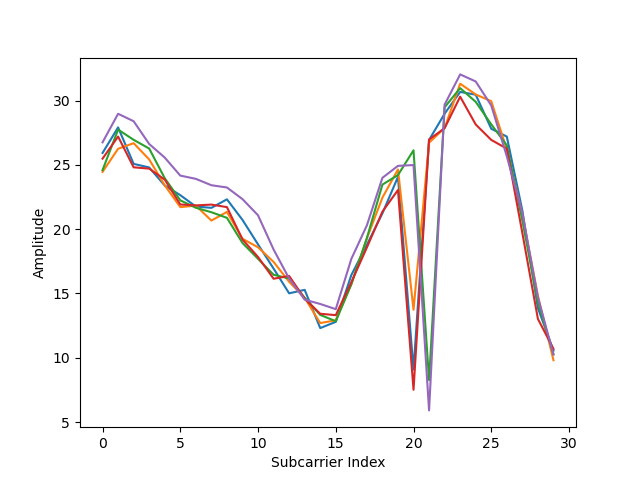}}}
\subfloat[\centering Original CSI Phase] {{\includegraphics[scale=0.35]{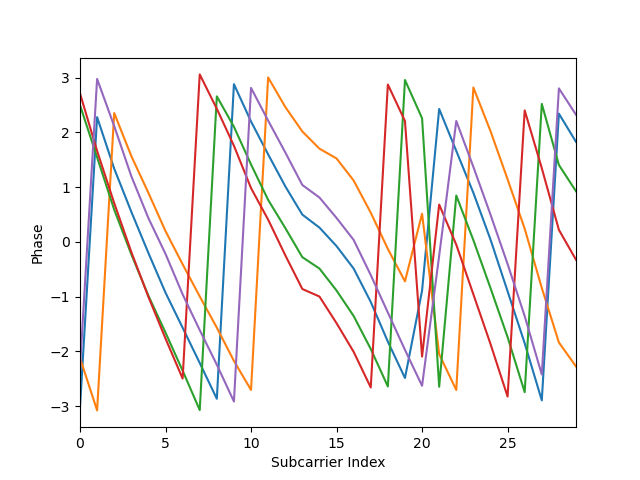} }}
\subfloat[\centering Phase after unwrapping ]{{\includegraphics[scale=0.35]{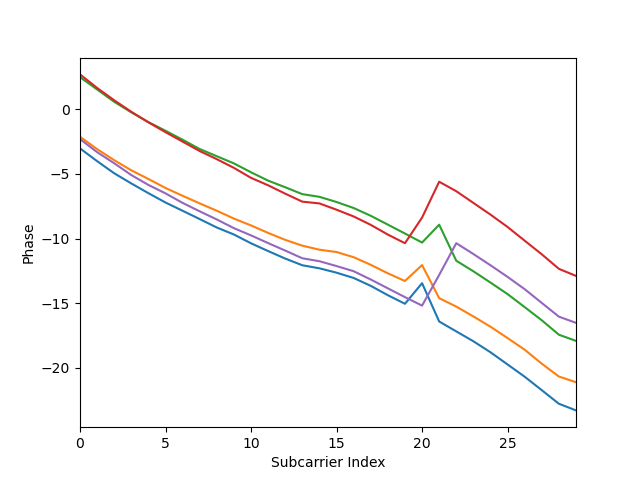}}} \\
\subfloat[\centering Phase after unwrapping + linear fitting ]{{\includegraphics[scale=0.35]{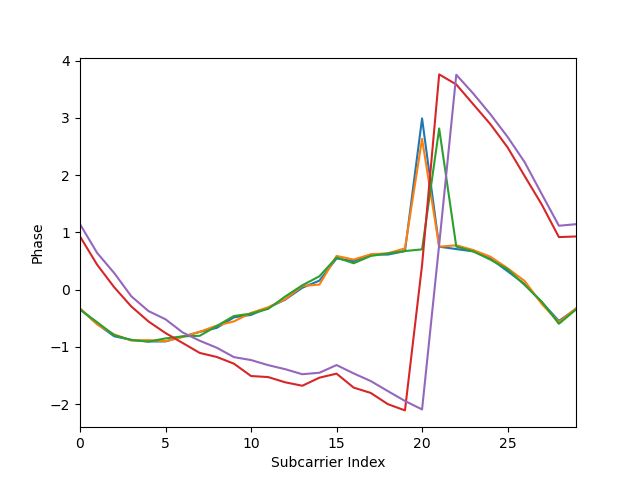}}}
\subfloat[\centering Phase after unwrapping + filtering] {{\includegraphics[scale=0.35]{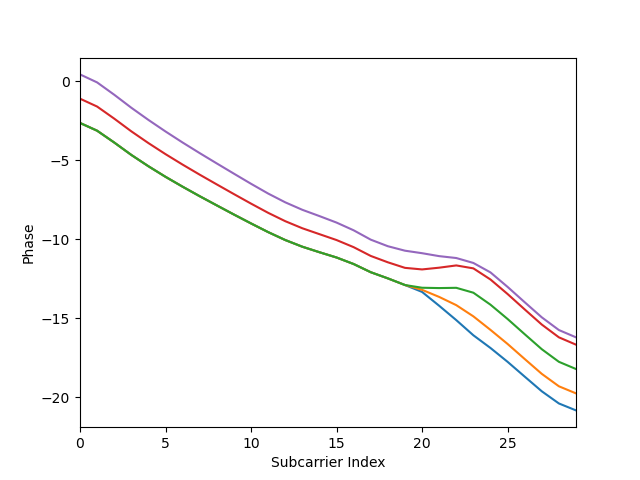} }}
\subfloat[\centering Phase after unwrapping + filtering + linear fitting] {{\includegraphics[scale=0.35]{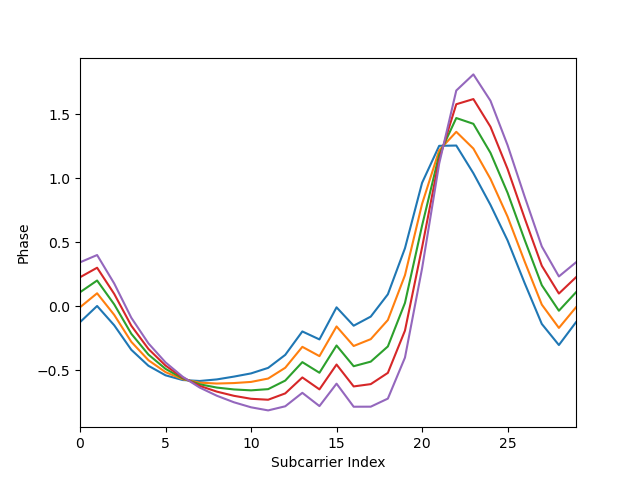}}}
\caption{Sanitization steps of CSI sequences described in Section 3.1. In each subfigure, we plot five consecutive samples (five colored curves) each containing CSI data of 30 IEEE 802.11n/ac sub-Carrier frequencies (horizontal axis).}
\label{fig:Sanitization}
\end{figure*}

The raw CSI samples are noisy with random phase drift and flip (see Figure~\ref{fig:Sanitization}(b)). Most WiFi-based solutions disregard the phase of CSI signals and rely only on their amplitude (see Figure~\ref{fig:Sanitization} (a)). As shown in our experimental validation, discarding the phase information have a negative impact on the performance of our model. In this section, we perform sanitization to obtain stable phase values to enable full use of the CSI information. 

In raw CSI samples (5 consecutive samples visualized in Figure~\ref{fig:Sanitization}(a-b)), the amplitude ($A$) and phase ($\Phi$) of each complex element $z=a+bi$ are computed using the formulation $A = \sqrt{ (a^2 + b^2)}$ and $\Phi = arctan(b/a)$. Note that the range of the $\arctan$ function is from $-\pi$ to $\pi$ and the phase values outside this range get wrapped, leading to a discontinuity in phase values. Our first sanitization step is to unwrap the phase following \cite{huawei-wifi}:
\begin{equation}
\begin{aligned}
\Delta \phi_{i,j} &= \Phi_{i,j+1} - \Phi_{i,j} \\
\text{if } \Delta \phi_{i,j} > \pi, \Phi_{i,j+1} &= \Phi_{i,j} + \Delta \phi_{i,j} - 2\pi\\
\text{if } \Delta \phi_{i,j} < -\pi, \Phi_{i,j+1} &= \Phi_{i,j} + \Delta \phi_{i,j} + 2\pi,\\
\end{aligned}
\end{equation}
where $i$ denotes the index of the measurements in the five consecutive samples, and $j$ denotes the index of the subcarriers(frequencies). Following unwrapping, each of the flipping phase curves in Figure~\ref{fig:Sanitization}(b) are restored to continuous curves in Figure~\ref{fig:Sanitization}(c). 

Observe that among the 5 phase curves captured in 5 consecutive samples in Figure~\ref{fig:Sanitization}(c), there are random jiterings that break the temporal order among the samples. To keep the temporal order of signals, previous work \cite{phase_sanitized} mentioned linear fitting as a popular approach. However, directly applying linear fitting to Figure~\ref{fig:Sanitization}(c) further amplified the jitering instead of fixing it (see the failed results in Figure~\ref{fig:Sanitization}(d)). 

\begin{figure*}[!htb]
\centering
\includegraphics[scale=0.13]{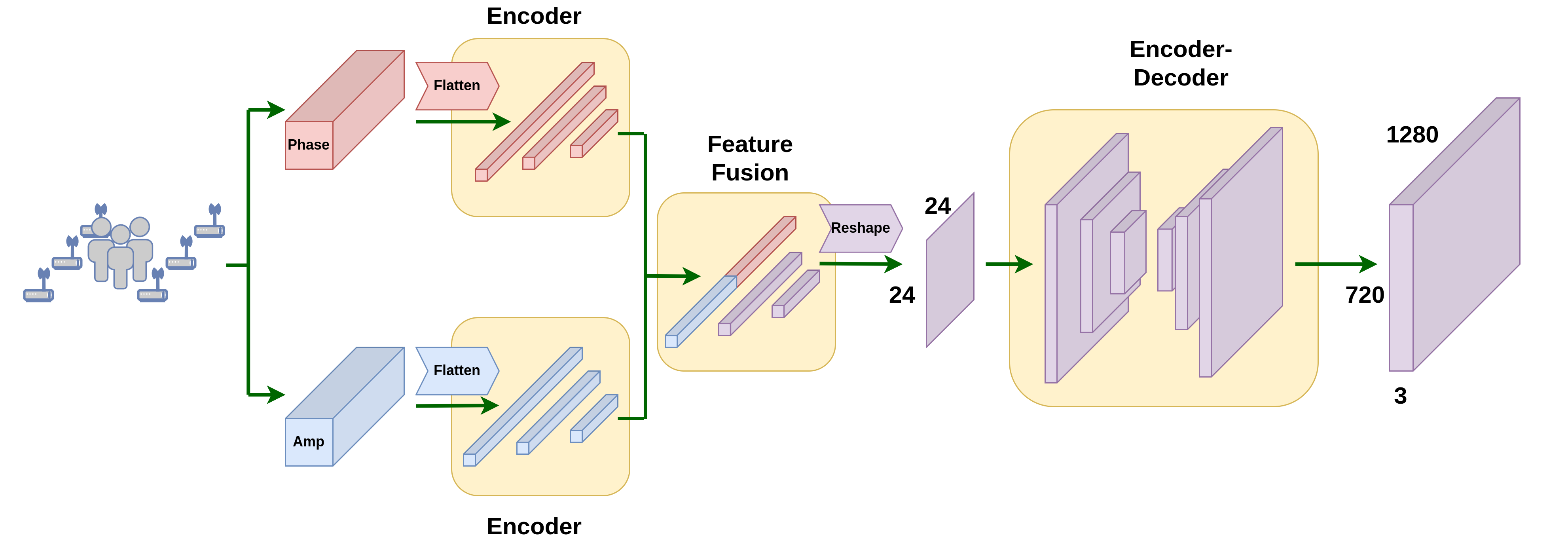}
\caption{Modality Translation Network. Two encoders extract the features from the amplitude and phase in the CSI domain. Then the features are fused and reshaped before going through an encoder-decoder network. The output is a $3 \times 720 \times 1280$ feature map in the image domain.} 
\label{fig:ModalityTranslation}
\end{figure*}

From Figure~\ref{fig:Sanitization}(c), we use median and uniform filters to eliminate outliers in both the time and frequency domain 
which leads to Figure~\ref{fig:Sanitization}(e). Finally, we obtain the fully sanitized phase values by applying the linear fitting method following the equations below:
\begin{equation}
\begin{aligned}
\alpha_1 &= \frac{\Phi_F - \Phi_1}{2\pi F}\\
\alpha_0 &= \frac{1}{F} \sum_{1\leq f \leq F} \phi_f\\
\hat{\phi_f} &= \phi_f - (\alpha_1 f + a_0),
\end{aligned}
\end{equation}
where $F$ denotes the largest subcarrier index (30 in our case) and $\hat{\phi_f}$ is the sanitized phase values at subcarrier $f$ (the $f$th frequency). In Figure~\ref{fig:Sanitization}(f), the final phase curves are temporally consistent.

 \subsection{Modality Translation Network}

In order to estimate the UV maps in the spatial domain from the 1D CSI signals, we first transform the network inputs from the CSI domain to the spatial domain. This is done with the Modality Translation Network (see Figure~\ref{fig:ModalityTranslation}). We first extract the CSI latent space features using two encoders, one for the amplitude tensor and the other for the phase tensor, where both tensors have the size of $150 \times 3 \times 3$ (5 consecutive samples, 30 frequencies, 3 emitters and 3 receivers). Previous work on human sensing with WiFi \cite{Person-in-wifi} stated that Convolutional Neural Network (CNN) can be used to extract spatial features from the last 
two dimensions (the $3 \times 3$ transmitting sensor pairs) of the input tensors. We, on the other hand, believe that locations in the $3 \times 3$ feature map do not correlate with the locations in the 2D scene. More specifically, as depicted in Figure~\ref{fig:CSIrepresenation}(b), the element that is colored in blue represents a 1D summary of the entire scene captured by emitter 1 and receiver 3 (E1 - R3), instead of local spatial information of the top right corner of the 2D scene. Therefore, we consider that each of the 1350 elements (in both tensors) captures a unique 1D summary
of the entire scene. Following this idea, the amplitude and phase tensors are flattened and feed into two separate multi-layer perceptrons (MLP) to obtain their features in the CSI latent space. We concatenated the 1D features from both encoding branches, then the combined tensor is fed to another MLP to perform feature fusion. 

The next step is to transform the CSI latent space features to feature maps in the spatial domain. As shown in Figure~\ref{fig:ModalityTranslation}, the fused 1D feature is reshaped into a $24 \times 24$ 2D feature map. Then, we extract the spatial information by applying two convolution blocks and obtain a more condensed map with the spatial dimension of $6 \times 6$. Finally, four deconvolution layers are used to upsample the encoded feature map in low dimensions to the size of $3 \times 720 \times 1280$. We set such an output tensor size to match the dimension commonly used in RGB-image-input network. We now have a scene representation in the image domain generated by WiFi signals.

\subsection{WiFi-DensePose RCNN}

After we obtain the $3 \times 720 \times 1280$ scene representation in the image domain, we can utilize image-based methods to predict the UV maps of human bodies. State-of-the-art pose estimation algorithms are two-stage; first, they run an independent person detector to estimate the bounding box and then conduct pose estimation from person-wise image patches. However, as stated before, each element in our CSI input tensors is a summary of the entire scene. It is not possible to extract the signals corresponding to a single person from a group of people in the scene. Therefore, we decide to adopt a network structure similar to DensePose-RCNN \cite{DensePose}, since it can predict the dense correspondence of multiple humans in an end-to-end fashion. 

\begin{figure*}[!htb]
\centering
\includegraphics[scale=0.12]{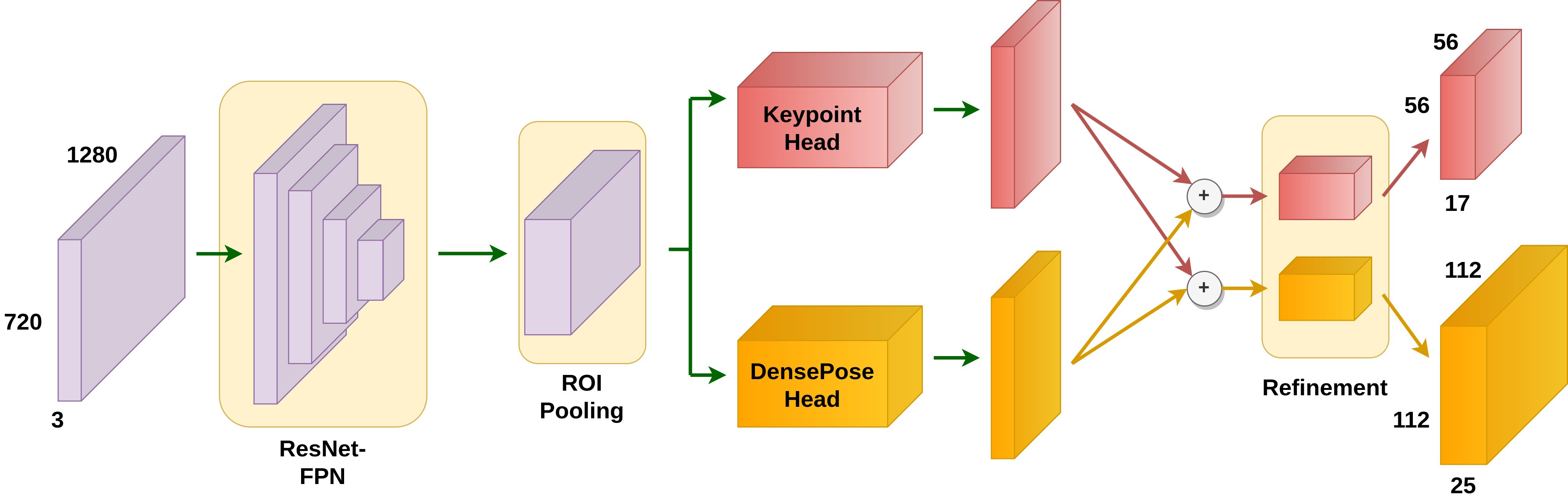}
\caption{WiFi-DensePose RCNN. The $3 \times 720 \times 1280$ feature map from Figure~\ref{fig:ModalityTranslation} first goes through standard ResNet-FPN and ROI pooling to extract person-wise features. The features are then processed by two heads:the Keypoint Head and the DensePose Head.} 
\label{fig:WiFi_DensePose_RCNN}
\end{figure*}

More specifically, in the WiFi-DensePose RCNN (Figure~\ref{fig:WiFi_DensePose_RCNN}), we extract the spatial features from the obtained $3 \times 720 \times 1280$ image-like feature map using the ResNet-FPN backbone~\cite{Resnet-fpn}. Then, the output will go through the region proposal network \cite{region_proposal}. To better exploit the complementary information of different sources, the next part of our network contains two branches: DensePose head and Keypoint head. Estimating keypoint locations is more reliable than estimating dense correspondences, so we can train our network to use keypoints to restrict DensePose predictions from getting too far from the body joints of humans. The DensePose head utilizes a Fully Convolutional Network (FCN) \cite{FCN} to densely predict human part labels and surface coordinates (UV coordinates) within each part, while the keypoint head uses FCN to estimate the keypoint heatmap. The results are combined and then fed into the refinement unit of each branch, where each refinement unit consists of two convolutional blocks followed by an FCN. The network outputs a $17 \times 56 \times 56$ keypoint mask and a $25 \times 112 \times 112$ IUV map. The process is demonstrated in Figure 5. It should be noted that the modality translation network and the WiFi-DensePose RCNN are trained together.

  
\subsection{Transfer Learning}
  
\begin{figure*}[!htb]
    \centering
    \includegraphics[scale=0.19]{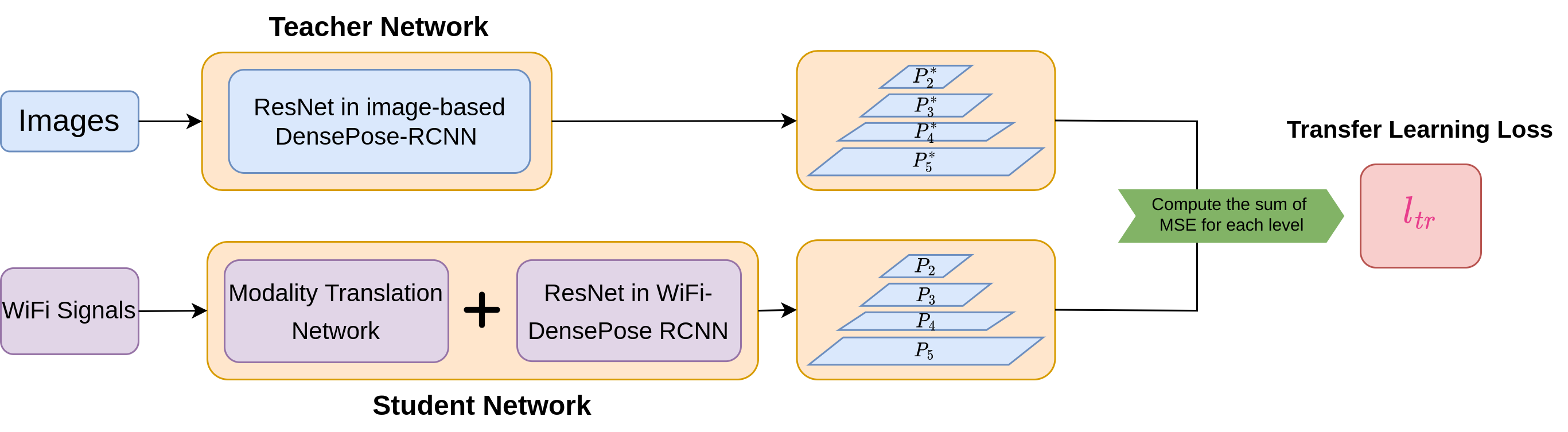} 
    \caption{Transfer learning from an image-based teacher network to our WiFi-based network.}
    
    \label{fig:TransferLearning}%
\end{figure*}
  
Training the Modality Translation Network and WiFi-DensePose RCNN network from a random initialization takes a lot of time (roughly 80 hours). To improve the training efficiency, we conduct transfer learning from an image-based DensPose network to our WiFi-based network (See Figure~\ref{fig:TransferLearning} for details). 
  
The idea is to supervise the training of the WiFi-based network with the pre-trained image-based network. Directly initializing the WiFi-based network with image-based network weights does not work because the two networks get inputs from different domains (image and channel state information). Instead, we first train an image-based DensePose-RCNN model as a teacher network.  Our student network consists of the modality translation network and the WiFi-DensePose RCNN. We fix the teacher network weights and train the student network by feeding them with the synchronized images and CSI tensors, respectively. We update the student network such that its backbone (ResNet) features mimic that of our teacher network. Our transfer learning goal is to minimize the differences of multiple levels of feature maps generated by the student model and those generated by the teacher model. Therefore we calculate the mean squared error between feature maps. The transfer learning loss from the teacher network to the student network is:
\begin{equation}
    L_{tr} = MSE(P_2, P^{*}_2) + MSE(P_3, P^{*}_3) + MSE(P_4 , P^{*}_4) + MSE(P_5, P^{*}_5),
\end{equation}
where $MSE(\cdot)$ computes the mean squared error between two feature maps, $\{P_2, P_3, P_4, P_5\}$ is a set of feature maps produced by the teacher network \cite{Resnet-fpn}, and $\{P^{*}_2, P^{*}_3, P^{*}_4, P^{*}_5\}$ is the set of feature maps produced by the student network \cite{Resnet-fpn}.

Benefiting from the additional supervision from the image-based model, the student network gets higher performance and takes fewer iterations to converge (Please see results in Table~\ref{tab:Ablation_1}).




  
  

\subsection{Losses}

The total loss of our approach is computed as:
\begin{eqnarray}
   \nonumber L &=& L_{cls} + L_{box} + \lambda_{dp} L_{dp} + \lambda_{kp} L_{kp}+ \lambda_{tr} L_{tr},
\end{eqnarray}
where $L_{cls}, L_{box}, L_{dp}, L_{kp}, L_{tr}$ are losses for the person classification, bounding box regression, DensePose, keypoints, and transfer learning respectively. The classification loss $L_{cls}$ and the box regression loss $L_{box}$ are standard RCNN losses \cite{Faster-RCNN,Mask-RCNN}.
The DensePose loss $L_{dp}$\cite{DensePose} consists of several sub-components: (1) Cross-entropy loss for the coarse segmentation tasks. Each pixel is classified as either belonging to the background or one of the 24 human body regions. (2) Cross-entropy loss for body part classification and smooth L1 loss for UV coordinate regression. These losses are used to determine the exact coordinates of the pixels, i.e., 24 regressors are created to break the full human into small parts and parameterize each piece using a local two-dimensional UV coordinate system, that identifies the position UV nodes on this surface part. 

We add $L_{kp}$ to help the DensePose to balance between the torso with more UV nodes and limbs with fewer UV nodes. Inspired by Keypoint RCNN \cite{Mask-RCNN}, we one-hot-encode each of the $17$ ground truth keypoints in one $56 \times 56$ heatmap, generating $17 \times 56 \times 56$ keypoints heatmaps and supervise the output with the Cross-Entropy Loss. To closely regularize the Densepose regression, the keypoint heatmap regressor takes the same input features used by the Denspose UV maps. 

\section{Experiments}

This section presents the experimental validation of our WiFi-based DensePose. 

\subsection{Dataset}
We used the dataset \footnote{The identifiable information in this dataset has been removed for any privacy concerns.} described in~\cite{Wanghuang2019}, which contains CSI samples taken at $100$Hz from receiver antennas and videos recorded at $20$ FPS. Time stamps are used to synchronize CSI and the video frames such that 5 CSI samples correspond to 1 video frame. The dataset was gathered in sixteen spatial layouts: six captures in the lab office and ten captures in the classroom. Each capture is around 13 minutes with 1 to 5 subjects (8 subjects in total for the entire dataset) performing daily activities under the layout described in Figure 2 (a). \textbf{The sixteen spatial layouts are different in their relative locations/orientations of the WiFi-emitter antennas, person, furniture, and WiFi-receiver antennas.}  

\begin{figure}[!htb]
\centering
\includegraphics[scale=0.2]{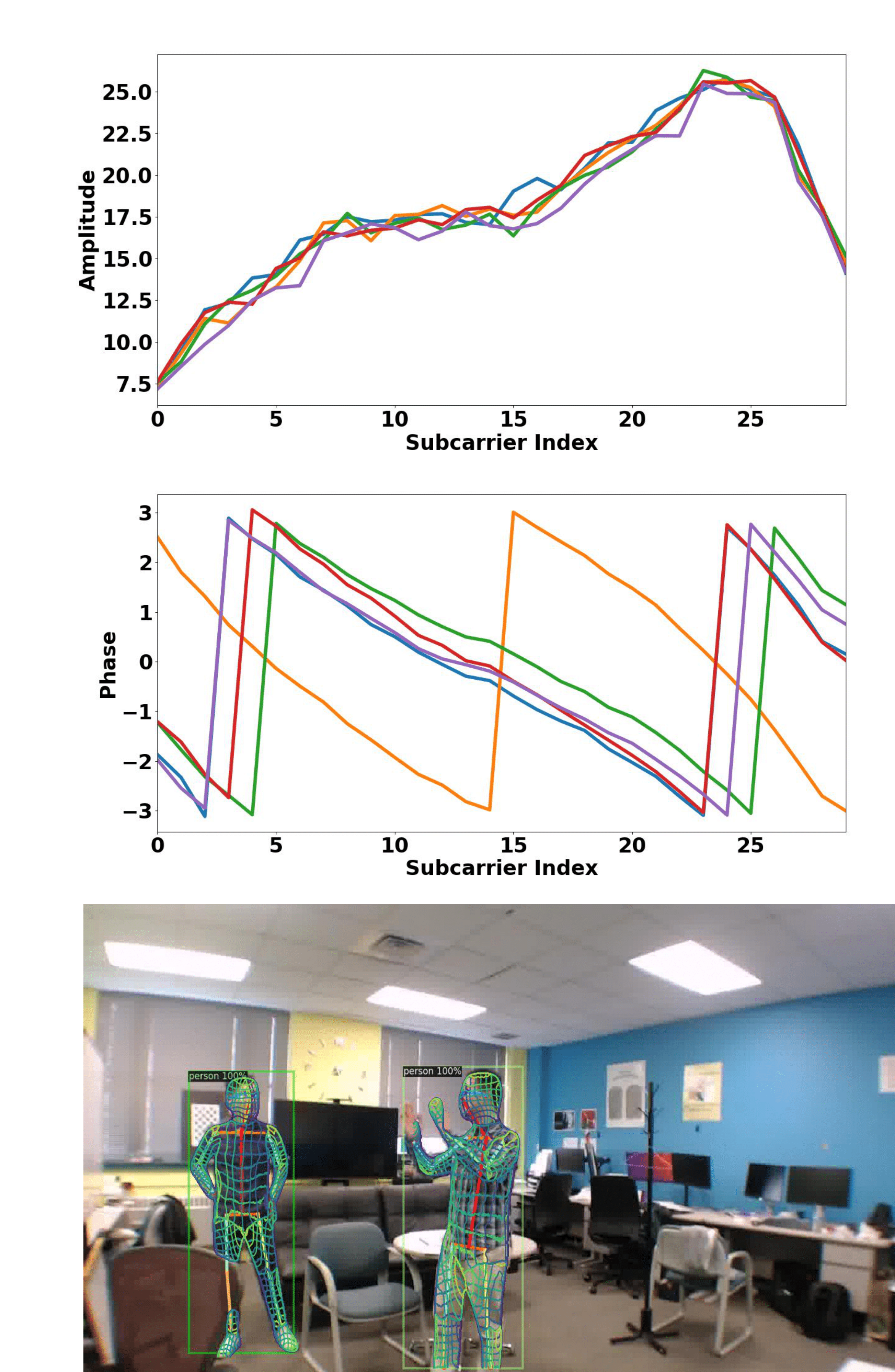}
\caption{Top two rows are the amplitude and phase of the input WiFi signal. The bottom row shows R101-Psuedo-GT: the ground truth dense pose and keypoints annotated by running a pretrained image-based Densepose network on the corresponding RGB image (see Section 4.1 for details).}
\label{fig:annotation}
\end{figure}

There are no manual annotations for the data set. We use the MS-COCO-pre-trained dense model "R\_101\_FPN\_s1x\_legacy" \footnote{\url{https://github.com/facebookresearch/detectron2/blob/main/projects/DensePose/doc/DENSEPOSE_IUV.md##ModelZoo}} and  MS-COCO-pre-trained Keypoint R-CNN "R101-FPN" \footnote{\url{https://github.com/facebookresearch/detectron2/blob/main/MODEL_ZOO.md##coco-person-keypoint-detection-baselines-with-keypoint-r-cnn}} to produce the pseudo ground truth. We denote the ground truth as "\textbf{R101-Pseudo-GT}" (see an annotated example in Figure~\ref{fig:annotation}). The R101-Pseudo-GT includes person bounding boxes, person-instance segmentation masks, body-part UV maps, and person-wise keypoint coordinates. 

Throughout the section, we use R101-Puedo-GT to train our WiFi-based DensePose model as well as finetuning the image-based baseline "R\_50\_FPN\_s1x\_legacy". 

\subsection{Training/Testing protocols and Metrics}
We report results on two protocols: (1) 
\textbf{Same layout}: We train on the training set in all $16$ spatial layouts, and test on remaining frames. Following \cite{Wanghuang2019}, we randomly select 80$\%$ of the samples to be our training set, and the rest to be our testing set. The training and testing samples are different in the person's location and pose, but share the same person's identities and background. This is a reasonable assumption since the WiFi device is usually installed in a fixed location. (2) \textbf{Different layout}: We train on 15 spatial layouts and test on 1 unseen spatial layout. The unseen layout is in the classroom scenarios.  

We evaluate the performance of our algorithm in two aspects: the ability to detect humans (bounding boxes)
and accuracy of the dense pose estimation. 

To evaluate the performance of our models in detecting humans, we calculate the standard average precision (AP) of person bounding boxes at multiple IOU thresholds ranging from 0.5 to 0.95.  



In addition, by MS-COCO \cite{coco} definition, we also compute AP-m for median bodies that are enclosed in bounding boxes with sizes between $32 \times 32$ and $96 \times 96$ pixels in a normalized $640 \times 480$ pixels image space, and AP-l for large bodies that are enclosed in bounding boxes larger than $96 \times 96$ pixels.



To measure the performance of DensePose detection, we follow the original DensePose paper \cite{DensePose}. We first compute Geodesic Point Similarity (GPS) as a matching score for dense correspondences:
\begin{equation}
GPS_j = \frac{1}{|P_j|}\sum_{p \in P_j}\exp(\frac{-g(i_p, \hat{i}_p)^2}{2\kappa^2}),
\end{equation}
where $g$ calculates the geodesic distance, $P_j$ denotes the ground truth point annotations of person $j$, $i_p$ and $\hat{i}_p$ are the estimated and ground truth vertex at point $p$ respectively, and $\kappa$ is a normalizing parameter (set to be 0.255 according to \cite{DensePose}). 

One issue of GPS is that it does not penalize spurious predictions. Therefore, estimations with all pixels classified as foreground are favored. To alleviate this issue, masked geodesic point similarity (GPSm) was introduced in \cite{DensePose}, which incorporates both GPS and segmentation masks. The formulation is as follows:
\begin{equation}
    GPSm = \sqrt{GPS\cdot I}, I = \frac{M \cap \hat{M}}{M \cup \hat{M}},
\end{equation}
where $M$ and $\hat{M}$ are the predicted and ground truth foreground segmentation masks. 

Next, we can calculate DensePose average precision with GPS (denoted as dpAP$\cdot$ GPS) and GPSm (denoted as dpAP$\cdot$ GPSm) as thresholds, following the same logic behind the calculation of bounding box AP. 




\subsection{Implementation Details}

We implemented our approach in PyTorch. We set the training batch size to 16 on a 4 GPU (Titan X) 
server. We empirically set $\lambda_{dp}=0.6, \lambda_{kp}=0.3, \lambda_{tr}=0.1$. We used a warmup multi-step learning rate scheduler and set the initial learning rate as $1e-5$. The learning rate increases to $1e-3$ during the first 2000 iterations, then decreases to $\frac{1}{10}$ of its value every 48000 iterations. We trained for $145,000$ iterations for our final model.



\subsection{WiFi-based DensePose under Same Layout}

Under the Same Layout protocol, we compute the AP of human bounding box detections as well as dpAP$\cdot$ GPS and dpAP$\cdot$ GPSm of dense correspondence predictions. Results are presented in Table~\ref{tab:WiFi-AP} and Table~\ref{tab:WiFi-DP-AP}, respectively.

\begin{table}[!htb]
\begin{center}
\begin{tabular}{ |c|c|c|c|c|c| } 
 \hline
 Method & AP & AP@50 & AP@75 & AP-m & AP-l \\ 
 \hline
 WiFi & 43.5 & 87.2 & 44.6 & 38.1 & 46.4\\ 
 \hline
\end{tabular}
\end{center}
\caption{Average precision (AP) of WiFi-based DensePose under the Same Layout protocol. All metrics are the higher the better.}
\label{tab:WiFi-AP}
\end{table}


From Table~\ref{tab:WiFi-AP}, we can observe a high value (87.2) of AP@50, indicating that our model can effectively detect the approximate locations of human bounding boxes. The relatively low value (35.6) for AP@75 suggests that the details of the human bodies are not perfectly estimated.

\begin{table*}[!htb]
\begin{center}
\begin{tabular}{ |c|c|c|c|c|c|c| } 
 \hline
 Method & dpAP $\cdot$ GPS & dpAP $\cdot$ GPS@50 & dpAP $\cdot$ GPS@75 & dpAP $\cdot$ GPSm & dpAP $\cdot$ GPSm@50 & dpAP $\cdot$ GPSm@75 \\ 
 \hline
 WiFi &  45.3 & 76.7 & 47.7 & 44.8 & 73.6 & 44.9\\ 
 \hline
\end{tabular}
\end{center}
\caption{DensePose Average precision (dpAP $\cdot$ GPS, dpAP $\cdot$ GPSm) of WiFi-based DensePose under the Same Layout protocol. All metrics are the higher the better.}
\label{tab:WiFi-DP-AP}
\end{table*}

A similar pattern can be seen from the results of DensePose estimations (see Table~\ref{tab:WiFi-DP-AP} for details). Experiments report high values of dpAP $\cdot$ GPS@50 and dpAP $\cdot$ GPSm@50, but low values of dpAP $\cdot$ GPS@75 and dpAP $\cdot$ GPSm@75. This demonstrates that our model performs well at estimating the poses of human torsos, but still struggles with detecting details like limbs.

\subsection{Comparison with Image-based DensePose}

\begin{table}[!htb]
\begin{center}
\begin{tabular}{ |c|c|c|c|c|c| } 
 \hline
 Method & AP & AP@50 & AP@75 & AP-m & AP-l \\ 
 \hline
 WiFi & 43.5 & 87.2 & 44.6 & 38.1 & 46.4\\ 
 \hline
 Image & 84.7 & 94.4 & 77.1 & 70.3 & 83.8\\ 
 \hline
\end{tabular}
\end{center}
\caption{Average precision (AP) of WiFi-based and image-based DensePose under the Same Layout protocol. All metrics are the higher the better.}
\label{tab:WiFi-Image-AP}
\end{table}

As discussed in Section 4.1, since there are no manual annotations on the WiFi dataset, it is difficult to compare the performance of WiFi-based DensePose with its Image-based counterpart. This is a common drawback of many WiFi perception works including~\cite{Wanghuang2019}.

Nevertheless, conducting such a comparison is still worthwhile in assessing the current limit of WiFi-based perception. We tried an image-based DensePose baseline "R\_50\_FPN\_s1x\_legacy"  finetuned on R101-Pseudo-GT under the Same Layout protocol. In addition, as shown in Figure~\ref{fig:Qualitative_1} and Figure~\ref{fig:Qualitative_2}, though certain defects still exist, the estimations from our WiFi-based model are reasonably well compared to the results produced by Image-based DensePose. 

In the quantitative results in Table~\ref{tab:WiFi-Image-AP} and Table~\ref{tab:WiFi-Image-DP-AP}, the image-based baseline produces very high APs due to the small difference between its ResNet50 backbone and the Resnet101 backbone used to generate R101-Pseudo-GT. This is to be expected. Our WiFi-based model has much lower absolute metrics. However, it can be observed from Table~\ref{tab:WiFi-Image-AP} that the difference between AP-m and AP-l values is relatively small for the WiFi-based model. We believe this is because individuals who are far away from cameras occupy less space in the image, which leads to less information about these subjects. On the contrary, WiFi signals incorporate all the information in the entire scene, regardless of the subjects' locations. 

\begin{table*}[!htb]
\begin{center}
\begin{tabular}{ |c|c|c|c|c|c|c| } 
 \hline
 Method & dpAP $\cdot$ GPS & dpAP $\cdot$ GPS@50 & dpAP $\cdot$ GPS@75 & dpAP $\cdot$ GPSm & dpAP $\cdot$ GPSm@50 & dpAP $\cdot$ GPSm@75 \\ 
 \hline
 WiFi &  45.3 & 79.3 & 47.7 & 43.2 & 77.4 & 45.5\\ \hline
 Image & 81.8 & 93.7 & 86.2 & 84.0 & 94.9 & 86.8\\
 \hline
\end{tabular}
\end{center}
\caption{DensePose Average precision (dpAP $\cdot$ GPS, dpAP $\cdot$ GPSm) of WiFi-based and image-based DensePose under the Same Layout protocol. All metrics are the higher the better.}
\label{tab:WiFi-Image-DP-AP}
\end{table*}

\begin{table*}[!htb]
\begin{center}
\begin{tabular}{ |c|c|c|c|c|c|c|c| } 
 \hline
 Method & AP & AP@50 & AP@75 & AP-m & AP-l & Number of Trained Iterations\\ 
 \hline
 Amplitude-only Model & 39.5 & 85.4 & 41.3 & 34.4 & 43.7 & 174000 \\
  \hline
 + Sanitized Phase Input & 40.3 & 85.9 & 41.9 & 34.6 & 44.5 & 180000 \\
 \hline
  + Keypoint Supervision & 42.9 & 86.8 & 44.1 &  38.0 & 45.8 & 186000 \\
 \hline
  + Transfer Learning & 43.5 & 87.2 & 44.6 & 38.1 & 46.4 & 145000\\ 
 \hline
\end{tabular}
\end{center}
\caption{Ablation study of human detection under the Same-layout protocol. All metrics are the higher the better.}
\label{tab:Ablation_1}
\end{table*}

\begin{table*}[!htb]
\begin{center}
\begin{tabular}{ |c|c|c|c|c|c|c| } 
 \hline
 Method & dpAP $\cdot$ GPS & dpAP $\cdot$ GPS@50 & dpAP $\cdot$ GPS@75 & dpAP $\cdot$ GPSm & dpAP $\cdot$ GPSm@50 & dpAP $\cdot$ GPSm@75\\ 
 \hline
 Amplitude-only Model & 40.6 & 76.6 & 41.5 & 39.7 & 75.1 & 40.3 \\
  \hline
 + Sanitized Phase Input & 41.2 & 77.4 & 42.3 & 40.1 & 75.7 & 40.5 \\
 \hline
  + Keypoint Supervision & 44.6 & 78.8 & 46.9 &  42.9 & 76.8 & 44.9 \\
 \hline
  + Transfer Learning & 45.3 & 79.3 & 47.7 & 43.2 & 77.4 & 45.5 \\ 
 \hline
\end{tabular}
\end{center}
\caption{Ablation study of DensePose estimation under the Same-layout protocol. All metrics are the higher the better.}
\label{tab:Ablation_2}
\end{table*}

\subsection{Ablation Study}
This section describes the ablation study to understand the effects of phase information, keypoint supervision, and transfer learning on estimating dense correspondences. Similar to section 4.4, the models analyzed in this section are all trained under the same-layout mentioned in section 4.2. 

We start by training a baseline WiFi model, which does not include the phase encoder, the keypoint detection branch, or transfer learning. The results are presented in the first row of both Table~\ref{tab:Ablation_1} and Table~\ref{tab:Ablation_2} as a reference.


{\bf Addition of Phase information:} We first examine whether the phase information can enhance the baseline performance. As shown in the second row of Table~\ref{tab:Ablation_1} and Table~\ref{tab:Ablation_2}, the results for all the metrics have slightly improved from the baseline. This proves our hypothesis that the phase can reveal relevant information about the dense human pose. 



{\bf Addition of a keypoint detection branch:} Having established the advantage of incorporating phase information, we now evaluate the effect of adding a keypoint branch to our model. The quantitative results are summarized in the third row of Table~\ref{tab:Ablation_1} and  Table~\ref{tab:Ablation_2}. 

Comparing with the numbers on the second row, we can observe a slight increase in performance in terms of dpAP$\cdot$GPS@50(from 77.4 to 78.8) and dpAP$\cdot$GPSm@50 (from 75.7 to 76.8), and a more noticeable improvement in terms of dpAP$\cdot$GPS@75 (from 42.3 to 46.9) and dpAP$\cdot$GPSm@75 (from 40.5 to 44.9). This indicates that the keypoint branch provides effective references to dense pose estimations, and our model becomes significantly better at detecting subtle details (such as the limbs).  


{\bf Effect of Transfer Learning:} We aim to reduce the training time for our model with the help of transfer learning. For each model in Table~\ref{tab:Ablation_1}, we continue training the model until there are no significant changes in terms of performance. The last row of Table~\ref{tab:Ablation_1} and Table~\ref{tab:Ablation_2} represents our final model with transfer learning. Though the final performance does not improve too much compared to the model (with phase information and keypoints) without transfer learning, it should be noted that the number of training iterations decreases significantly from 186000 to 145000 (this number includes the time to perform transfer learning as well as training the main model). 


\subsection{Performance in different layouts}

\begin{table*}[!htb]
\begin{center}
\begin{tabular}{ |c|c|c|c|c|c| } 
 \hline
 Method & AP & AP@50 & AP@75 & AP-m & AP-l \\
 \hline
 WiFi (base) & 23.5 & 48.1 & 20.3 & 19.4 & 24.5\\
 \hline
 WiFi (final) & 27.3 & 51.8 & 24.2 & 22.1 & 28.6\\
 \hline
 Image & 60.6 & 80.4 & 52.1 & 48.3 & 65.8\\ 
 \hline
\end{tabular}
\end{center}
\caption{Average precision (AP) of WiFi-based and image-based DensePose under the Different Layout protocol. All metrics are the higher the better.}
\label{tab:different_1}
\end{table*}

\begin{table*}[!htb]
\begin{center}
\begin{tabular}{ |c|c|c|c|c|c|c| } 
 \hline
 Method & dpAP $\cdot$ GPS & dpAP $\cdot$ GPS@50 & dpAP $\cdot$ GPS@75 & dpAP $\cdot$ GPSm & dpAP $\cdot$ GPSm@50 & dpAP $\cdot$ GPSm@75 \\
  \hline
 WiFi (base) &  22.3 & 47.3 & 21.5 & 20.9 & 44.6 & 21.8\\
 \hline
 WiFi (final) &  25.4 & 50.2 & 24.7 & 23.2 & 47.4 & 26.5\\ \hline
 Image & 60.2 & 70.1 & 62.3 & 54.0 & 72.7 & 58.8\\
 \hline
\end{tabular}
\end{center}
\caption{DensePose Average precision (dpAP $\cdot$ GPS, dpAP $\cdot$ GPSm) of WiFi-based and image-based DensePose under the Different Layout protocol. All metrics are the higher the better.}
\label{tab:different_2}
\end{table*}

\begin{figure*}
\subfloat[\centering ]{{\includegraphics[scale=0.18]{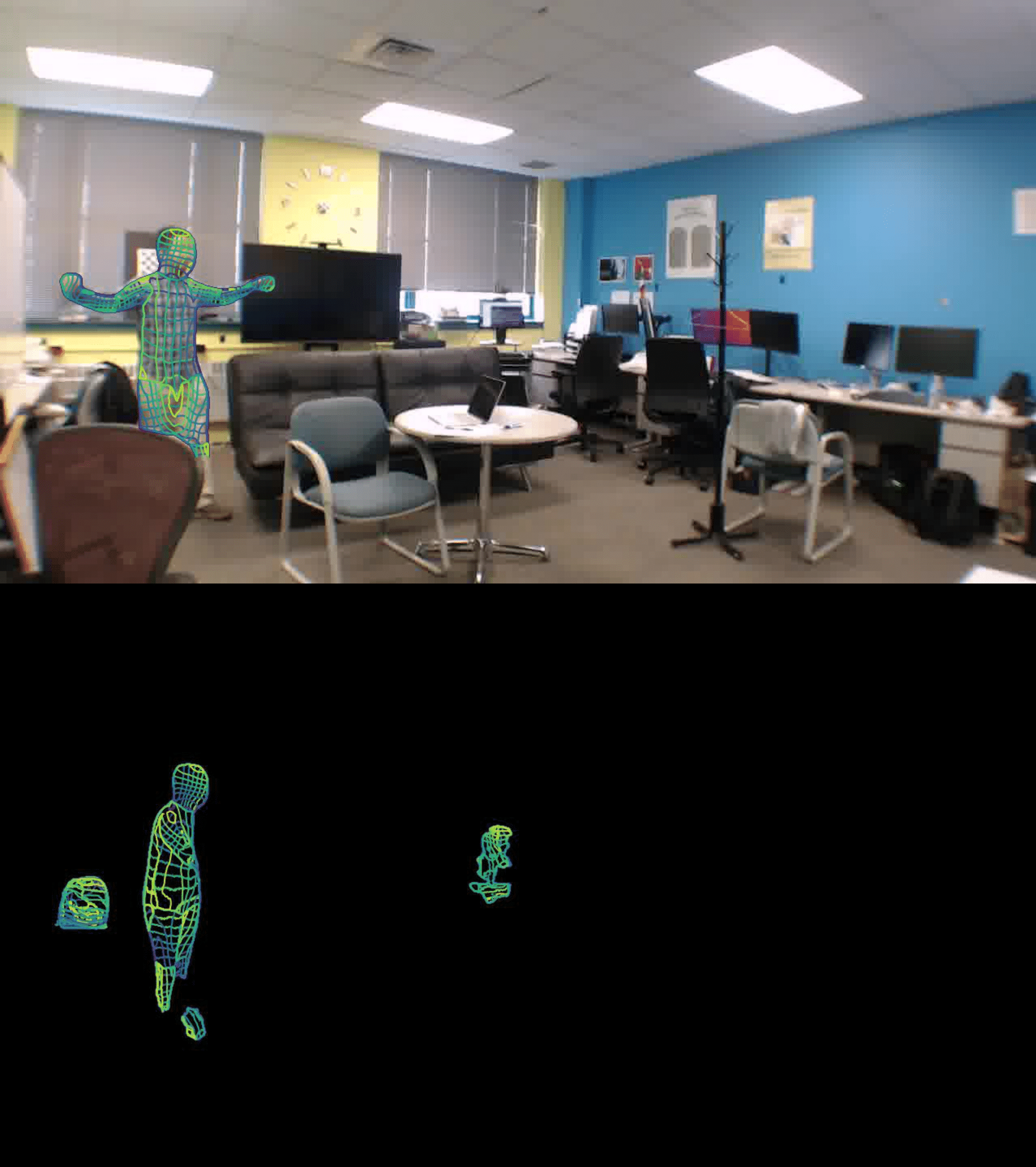}}}
\subfloat[\centering]{{\includegraphics[scale=0.18]{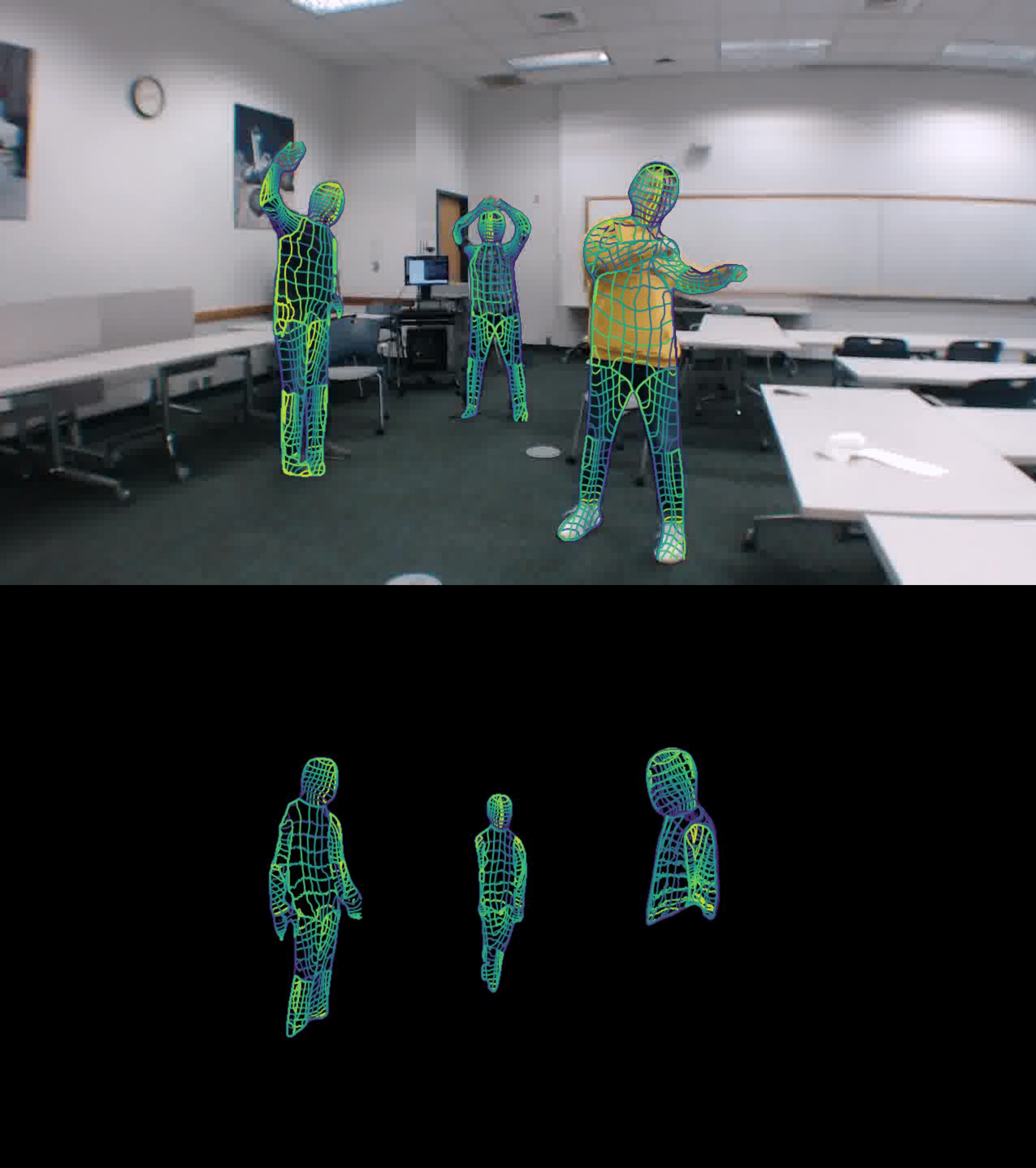}}}
\subfloat[\centering]{{\includegraphics[scale=0.18]{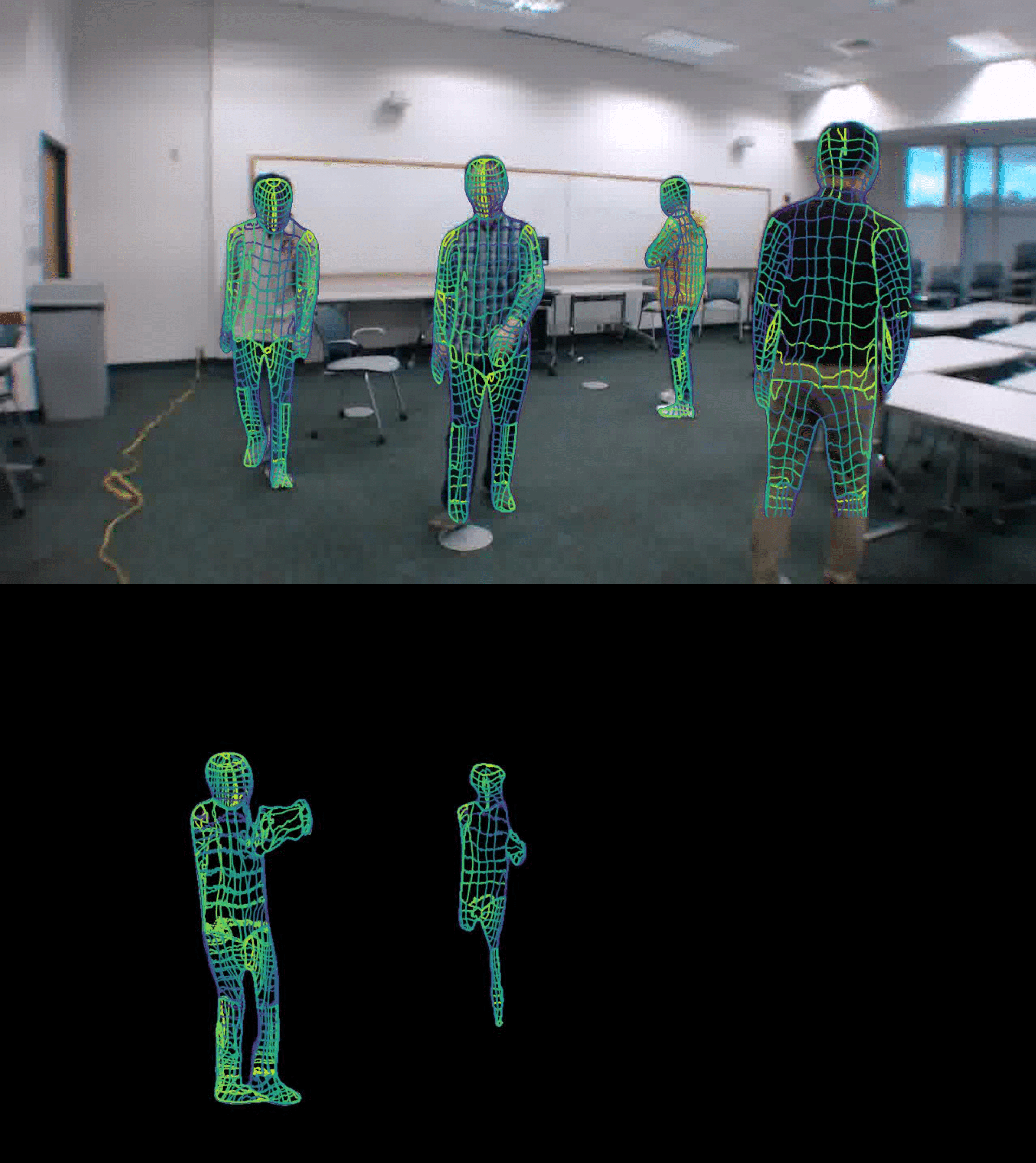}}}
\subfloat[\centering]{{\includegraphics[scale=0.18]{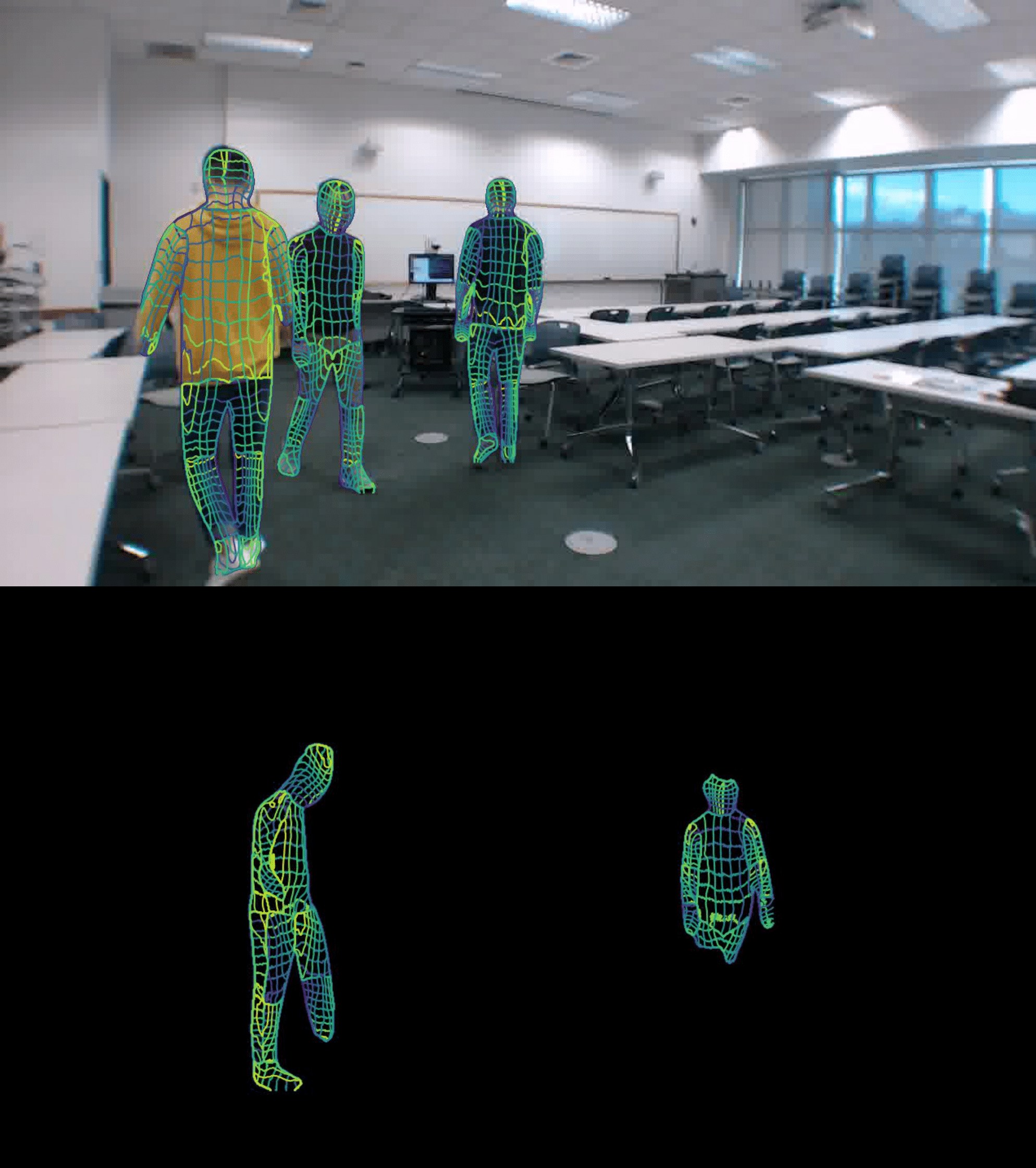}}}
\caption{Examples pf failure cases: (a-b) rare poses; (c-d) Three or more concurrent subjects. The first row is ground truth dense pose estimation. The second row illustrates the predicted dense pose.}
\label{fig:Fail}
\end{figure*}

All above results are obtained using the same layout for training and testing. However, WiFi signals in different environments exhibit significantly different propagation patterns. Therefore, it is still a very challenging problem to deploy our model on data from an untrained layout.  

To test the robustness of our model, we conducted the previous experiment under the different layout protocols, where there are 15 training layouts and 1 testing layout. The experimental results are recorded in Table~\ref{tab:different_1} and Table~\ref{tab:different_2}.

We can observe that our final model performs better than the baseline model in the unseen domain, but the performance decreases significantly from that under the same layout protocol: the AP performance drops from $43.5$ to $27.3$ and dpAP$\cdot$GPS drops from $45.3$ to $25.4$. However, it should also be noted that the image-based model suffers from the same domain generalization problem. We believe a more comprehensive dataset from a wide range of settings can alleviate this issue. 


\begin{figure*}[!htb]
\centering
\includegraphics[scale=0.48]{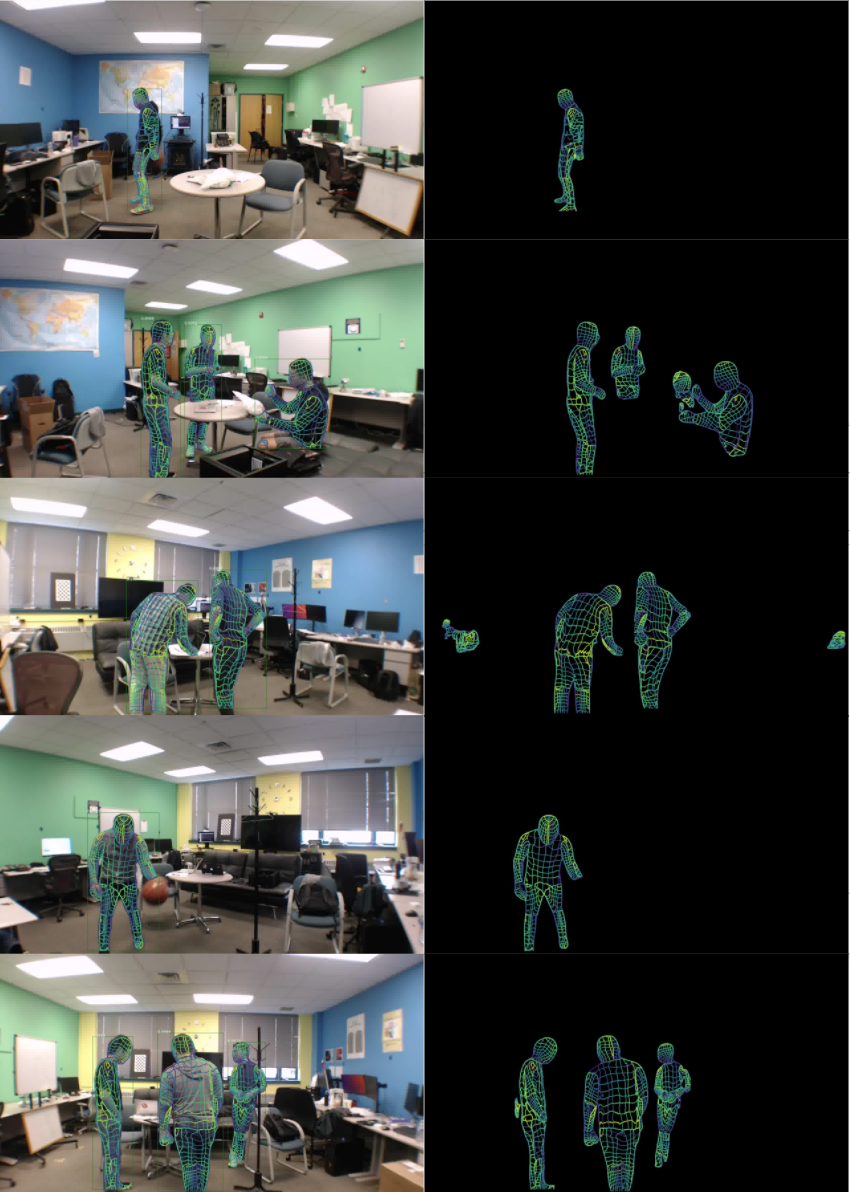}
\caption{Qualitative comparison using synchronized images and WiFi signals. (Left Column) image-based DensePose (Right Column) our WiFi-based DensePose.}
    \label{fig:Qualitative_1}
\end{figure*}

\begin{figure*}[!htb]
\centering
\includegraphics[scale=0.45]{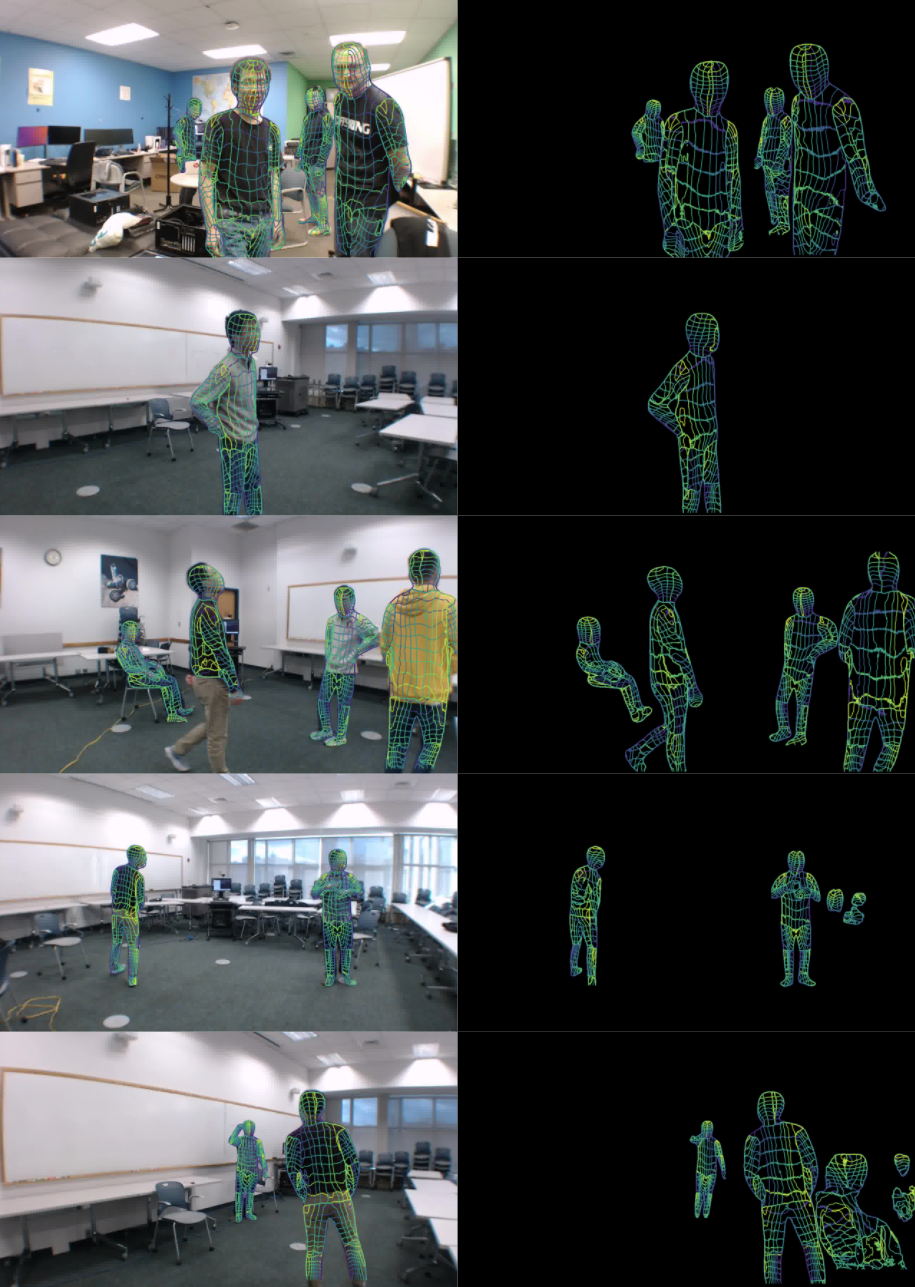}
\caption{More qualitative comparison using synchronized images and WiFi signals. (Left Column) image-based DensePose (Right Column) our WiFi-based DensePose.}
    \label{fig:Qualitative_2}
\end{figure*}

\subsection{Failure cases}
We observed two main types of failure cases. (1) When there are body poses that rarely occurred in the training set, the WiFi-based model is biased and is likely to produce wrong body parts (See examples (a-b) in Figure~\ref{fig:Fail}). (2) When there are three or more concurrent subjects in one capture, it is more challenging for the WiFi-based model to extract detailed information for each individual from the amplitude and phase tensors of the entire capture. (See examples (c-d) in Figure~\ref{fig:Fail}). We believe both of these issues can be resolved by obtaining more comprehensive training data.


\section{Conclusion and future work}
In this paper, we demonstrated that it is possible to obtain dense human body poses from WiFi signals by utilizing deep learning architectures commonly used in computer vision. Instead of directly training a randomly initialized WiFi-based model, we explored rich supervision information to improve both the performance and training efficiency, such as utilizing the CSI phase, adding keypoint detection branch, and transfer learning from an image-based model. The performance of our work is still limited by the public training data in the field of WiFi-based perception, especially under different layouts. In future work, we also plan to collect multi-layout data and extend our work to predict 3D human body shapes from WiFi signals. We believe that the advanced capability of dense perception could empower the WiFi device as a privacy-friendly, illumination-invariant, and cheap human sensor compared to RGB cameras and Lidars. 

\bibliographystyle{ACM-Reference-Format}
\bibliography{densepose_ref}

\end{document}